%% file: main.tex
\documentclass[conference]{IEEEtran}
\IEEEoverridecommandlockouts

\usepackage{fancyhdr}
\usepackage{cite}
\usepackage{amsmath,amssymb,amsfonts}
\usepackage{algorithmic}
\usepackage{graphicx}
\usepackage{textcomp}
\usepackage{xcolor}
\usepackage{url}
\usepackage{hyperref}
\usepackage{ifthen}
\usepackage{multirow}
\usepackage{makecell}
\usepackage[normalem]{ulem}

\def\BibTeX{{\rm B\kern-.05em{\sc i\kern-.025em b}\kern-.08em
    T\kern-.1667em\lower.7ex\hbox{E}\kern-.125emX}}

\newboolean{finalversion}
\setboolean{finalversion}{true} 

\newcommand{\statusTodo}{todo}
\newcommand{\statusDoing}{doing}
\newcommand{\statusDone}{done}

\newcommand{\TODO}[2][\statusTodo]{%
    \ifthenelse{\boolean{finalversion}}%
    {}
    {%
        \ifthenelse{\equal{#1}{\statusTodo}}%
        {{\color{red}[TODO] #2}}
        {%
            \ifthenelse{\equal{#1}{\statusDoing}}%
            {{\color{green}[DOING] #2}}
            {%
                \ifthenelse{\equal{#1}{\statusDone}}%
                {\sout{[DONE] #2}}
                {{\color{red}[TODO] #2}}
            }%
        }%
    }%
}

\fancypagestyle{firstpage}{
  \fancyhf{} 
  \fancyfoot[L]{\footnotesize
    © 2024 IEEE. Personal use of this material is permitted. Permission from IEEE must be
    obtained for all other uses, in any current or future media, including
    reprinting/republishing this material for advertising or promotional purposes, creating new
    collective works, for resale or redistribution to servers or lists, or reuse of any copyrighted
    component of this work in other works.
  }
}

\begin{document}

\title{Non-Uniform Parameter-Wise Model Merging
\thanks{
This work was partially funded by the Natural Sciences and Engineering Research Council of Canada (NSERC) CGS D 569345 - 2022 scholarship [S.H.]; FRQNT-NSERC grant 2023-NOVA-329125 [E.B. \& G.W.]; Canada CIFAR AI Chair, NSF DMS grant 2327211 and NSERC Discovery grant 03267 [G.W.]. This work is also supported by resources from Compute Canada and Calcul Quebec. The content is solely the responsibility of the authors and does not necessarily represent the views of the funding agencies.}
}

\newcommand{\concordia}[1]{
\IEEEauthorblockA{
\textit{Concordia University} \\
\textit{Mila - Québec AI Insitute} \\
Montréal, Canada \\
\url{#1}
}}

\newcommand{\udem}[1]{
\IEEEauthorblockA{
\textit{Université de Montréal} \\
\textit{Mila - Québec AI Insitute} \\
Montréal, Canada \\
\url{#1}
}}

\author{\IEEEauthorblockN{Albert M. Orozco Camacho}
\concordia{alorozco53@mila.quebec}
\and
\IEEEauthorblockN{Stefan Horoi}
\udem{stefan.horoi@umontreal.ca}
\and
\IEEEauthorblockN{Guy Wolf}
\udem{guy.wolf@umontreal.ca}
\and
\IEEEauthorblockN{Eugene Belilovsky}
\concordia{eugene.belilovsky@concordia.ca}
}


\maketitle
\thispagestyle{firstpage}


\begin{abstract}
Combining multiple machine learning models has long been a technique for enhancing performance, particularly in distributed settings. Traditional approaches, such as model ensembles, work well, but are expensive in terms of memory and compute. Recently, methods based on averaging model parameters have achieved good results in some settings and have gained popularity. 
However, merging models initialized differently that do not share a part of their training trajectories can yield worse results than simply using the base models, even after aligning their neurons.
In this paper, we introduce a novel approach, Non-uniform Parameter-wise Model Merging, or NP Merge, which merges models by learning the contribution of each parameter to the final model using gradient-based optimization. We empirically demonstrate the effectiveness of our method for merging models of various architectures in multiple settings, outperforming past methods.
We also extend NP Merge to handle the merging of multiple models, showcasing its scalability and robustness.
\end{abstract}


\input{sections/introduction}
\input{sections/related_work}
\input{sections/methods}
\input{sections/results}
\input{sections/discussion_and_conclusion}



\bibliographystyle{IEEEtran}
\bibliography{bibliography}

\end{document}

%% file: sections/introduction.tex
\section{Introduction}
\label{sec:intro}

Combining multiple machine learning (ML) models has been historically a popular approach to improve model performance.
Early works on model ensembles have shown that combining multiple models can lead to better
generalization and robustness \cite{ho1995random, lobacheva2020power}; in those settings, the focus has been on combining task predictions rather than the models themselves. 
On the other hand, combining model parameters brings the advantage of reducing the computational and memory overhead of inference and storage.
Furthermore, merging the models themselves is critical in federated learning (FL) settings to increase data privacy. In such settings, multiple models are trained independently on separate data instances that cannot be shared. Model ensembles are therefore not an option since this constraint prevents us from computing the predictions of the different models to then combine them.
Directly merging the models' parameters, which can be shared, offers another way of combining the learnings from the different models. This parameter agregation has come to be known as \textbf{model fusion} or \textbf{model merging}, and previous works have demonstrated its efficacy in many settings \cite{mcmahan17_fedavg, singh-jaggi2020_OT-fusion, tatro2020_opt-mode_con,ainsworth2023_git-rebasin, pena2023_sinkhorn-rebasin, jordan2023repair, stoica2023zipit, wortsman22_modelsoups, ilharco2023_task-arithmetic, wortsman2022_robust_finetuning, matena_raffel2022_fisher_weighted_averaging, jolicoeurmartineau2023_papa}. However, combining multiple models into a single one through model fusion is not a trivial task, and it is often difficult to achieve good predictive performance in this manner.

In the context of model merging, \emph{linear mode connectivity} (LMC) is the property of two local minima of a deep learning (DL) model which are connected by a linear low loss path in the parameter space of the model. This is a desirable property since LMC between two models ensures that we can simply linearly interpolate between their parameters without significantly impacting performance. However, linear mode connectivity is hard to achieve and isn't guaranteed even for models trained on the same dataset and with a shared random initialization \cite{pmlr-v119-frankle20a_lmc_lth}.
\cite{entezari2022_perm-invariance-lmc} recently conjectured that this difficulty in achieving LMC between models is largely due to the permutation invariance of deep learning models, i.e. the fact that we can change the order of the neurons in a given layer and the modified model would still represent the same mathematical function as long as the incoming and outgoing weights were permuted accordingly. Therefore most local minima of a given model, found through stochastic gradient descent, can become linearly mode-connected simply by finding the right permutations to align one model's parameters to those of the second model. Recent works have since proposed numerous methods for finding adequate permutations or linear transformations that align one model, say model $A$, to another, model $B$ \cite{tatro2020_opt-mode_con, singh-jaggi2020_OT-fusion, pena2023_sinkhorn-rebasin, ainsworth2023_git-rebasin, jordan2023repair, horoi2024harmony}. After aligning these models, their parameters can be combined through the equation $\alpha\mathcal{P}^A+(1-\alpha)\mathcal{P}^B$ where $\alpha\in(0,1)$ and $\mathcal{P}^A$ (resp. $\mathcal{P}^B$) are the parameters of model $A$ (resp. $B)$. These approaches have successfully reduced the loss barrier on the linear path between different models. 
For most models however, a loss in accuracy is still incurred when the parameters of the models being merged are linearly combined uniformly, i.e. $\alpha$ is a scalar that multiplies all the parameters in model $A$ and $B$ (through $1-\alpha$) uniformly. Recently, \cite{dhawan2024fedfish} argued that direct parameter averaging might fail to represent well either of the two mathematical functions learned by the two models being merged. They then propose a weighted merging of the models' parameters where the models' Fisher information determines the weights, a similar idea was presented in \cite{matena_raffel2022_fisher_weighted_averaging}


We continue this line of work, highlighting that while permutations might allow the alignment of different models in a way that makes LMC achievable, not all parameters are created equal, therefore assigning the same weight to all parameters of a model in the merging might not be the best way to take advantage of this newfound LMC between the models. Instead,
we start from permutation-based model alignment which helps attain the LMC property, as suggested by past works, and propose a novel approach to aggregate the models' parameters by learning each parameter's contribution to the final model through gradient-based optimization. Our approach, Non-uniform Parameter-wise Model Merging, or NP Merge, is more flexible than past approaches and thus leads to better-performing merged models with minimal additional computing. 


NP Merge’s allows for more precise control over how each parameter is interpolated, offering greater flexibility in merging models trained on diverse or non-overlapping datasets. By learning interpolation coefficients at the parameter level, NP Merge can adapt more effectively to the unique characteristics of each model, improving performance even in difficult merging scenarios. This flexibility ensures that the merged model retains high accuracy and generalization capability without being constrained by uniform aggregation strategies.
Furthermore, this line of work is orthogonal to some of the past works which have mainly focused on finding the right permutations to align different models but then all aggregate the parameters in the same way, through a uniform linear combination. 
NP Merge can therefore be used in conjunction with any of the previously proposed model alignment methods.

\paragraph*{\textbf{Contributions}}

Concretely, we make the following contributions:
\begin{itemize}
\item We propose \textit{Non-uniform Parameter-wise Model Merging}, or \textbf{NP Merge}, a robust approach to aggregate the parameters of multiple models, which were already previously aligned by existing methods, by learning the contribution of each parameter to the final model (Sec. \ref{ss:npmerge}). Rather than combining the parameters through a uniform linear combination, our method learns the contribution of each individual parameter to the final merged model through gradient-based optimization.
\item We empirically demonstrate the effectiveness of NP Merge when merging two models of various architectures trained in a multitude of settings, from using the same training data to training on completely disjoint subsets of the classes (Sec. \ref{ss:same_distribution} and Sec. \ref{ss:results-nonuniform}); when models come from the same initialization, 
we show that our approach approximates individual models better than previous works do.
\item For data-dependent merging methods, we analyze the impact of the amount of available data and show that NP-Merge possesses greater performance stability when lesser data is available.
\item We exhaustively assess NP Merge's performance against fine-tuning a merged model over several epochs. In particular, we show when each approach is suitable depending on diverse data distributions. Furthermore, we argue that using either approach brings more generalization benefits than any particular merging approach.
\item We extend our method to be able to merge multiple models through repetitive pairwise merging and show that this approach outperforms known state-of-the-art (SOTA) methods \ref{ss:many_models}.
\end{itemize}

%% file: sections/related_work.tex
\section{Related Work}
\label{sec:rw}

Artificial neural networks have highly complex and non-convex loss landscapes. How we can train such models despite this fact has puzzled researchers for a long time. Furthermore, ANNs are often overparameterized for the tasks they solve and they can memorize the tasks they are trained on \cite{arpit2017_memorization}, however, they're still able to generalize and learn important patterns from the data.

Multiple works have looked at the geometry of ANNs' loss landscapes \cite{Goldstein_LL_NIPS2018_7875, horoi2022_exploring}. Of particular interest here, \cite{freeman2017topology} theoretically showed that minima of one-layered ReLU networks become connected by low loss paths in parameter space as the number of neurons increases, i.e. they have asymptotically connected level sets. \cite{garipov2018_fge} and \cite{draxler2018_no-barriers} have continued this line of work empirically by providing algorithms to find non-linear low-loss paths between minima in ANNs' loss landscapes, this property of minima connected by such paths has been called \emph{mode connectivity}. \cite{garipov2018_fge} used these insights to propose Fast Geometric Ensembling (FGE) where multiple local minima from a single training trajectory with cyclical learning rates are ensembled to improve performance. \cite{pmlr-v119-frankle20a_lmc_lth} have introduced the term \emph{linear mode connectivity} (LMC) to describe the property of ANN minima connected by a linear low loss path in parameter space and have used it to study ANNs' stability to SGD noise, i.e. different data orders and augmentations. Specifically, they found that even models trained from the same initialization but with different SGD noise might not be linearly mode connected, 
however, sharing a part of the training trajectory helps models achieve LMC.
We note that the concept of LMC is important here since it ensures that the models are already \emph{aligned} to some extent, the $\mathtt{Align}$ function from Sec. \ref{sec:methods} can then simply be the identity and the aggregation of the models parameters ($\mathtt{Agg}$ function from Sec. \ref{sec:methods}) can consist of linearly interpolating the models' parameters.

\cite{entezari2022_perm-invariance-lmc} conjectured that ``Most SGD solutions belong to a set $\mathcal{S}$ whose elements can be permuted in such a way that there is no barrier on the linear interpolation between any two permuted elements in $\mathcal{S}$.'' In other words, most trained models found through standard ANN training are linearly mode connected, provided the right permutation is applied to align the two models' parameters. This conjecture is justified by the fact that one can apply any permutation to an ANN layer's neurons and their connections and the mathematical function described by the network will not have changed \cite{chen1993_geometryfeedforward, hechtnielsen1990_algebraicstructure}.

\paragraph{Works focused on model alignment} Numerous recent works have been focused on finding these ``right'' permutations to align different networks.
\cite{tatro2020_opt-mode_con} aligns networks by first computing the cross-correlation matrix between the neural activations of the two models and then solving the assignment problem maximizing the correlations of aligned neurons.
\cite{singh-jaggi2020_OT-fusion} uses optimal transport methods to find the permutations best matching neural activations or neural weights, with the method extending beyond permutations in the case where the number of neurons is not the same in the two models.
\cite{ainsworth2023_git-rebasin} proposes an iterative algorithm to align two models based on solving the assignment problem of minimizing the distance between the models' weights.
\cite{pena2023_sinkhorn-rebasin} proposes a differentiable algorithm that finds soft permutation matrices to align model neurons through the Sinkhorn Operator.
\cite{jordan2023repair} observed that model merging often leads to variance collapse in the merged model's activations and proposes REPAIR, a method to restore the internal statistics of averaged neural networks.
\cite{stoica2023zipit} goes beyond simply aligning models since they compute the correlations between the neurons of all models being merged and combine their parameters, starting with the highest correlated neurons, until the desired layer size is obtained. 

The works presented in this last paragraph are mainly concerned with aligning models ($\mathtt{Align}$ operation of Sec. \ref{sec:methods}) and pay little to no attention to the aggregation of the model's parameters ($\mathtt{Agg}$ operation of Sec. \ref{sec:methods}) which is done following Eq. \ref{eq:model_interpolation} with a single scalar $\alpha\in(0,1)$ for the whole model. Either multiple values for $\alpha$ in the $(0,1)$ range are sampled and the value with the highest loss (resp. lowest accuracy) is reported, i.e. the \emph{loss barrier} (resp. \emph{accuracy barrier}), or simply the loss/accuracy of the models average ($\alpha=0.5$) is reported.

\paragraph{Scenarios where models do not require alignment}
\cite{pmlr-v119-frankle20a_lmc_lth} has established that models often become stable to SGD noise early in training, meaning that models sharing a part of their training trajectory are often linearly mode connected. Along the same lines \cite{neyshabur2020_transfer} showed that when training from pre-trained weights, the models stay in the same loss landscape basin and therefore achieve LMC. In such cases, the models are already aligned and the $\mathtt{Align}$ operation can simply be the identity.
Numerous works have taken advantage of this phenomenon, for example, \cite{szegedy2016_inception} used a running average of the parameters found during training to evaluate their models, \cite{izmailov2018_averaging} found that simple averaging of multiple points along SGD trajectories leads to better generalization. \cite{wortsman2022_robust_finetuning} found that linearly interpolating the parameters of a pre-trained model and its fine-tuned version yields models that are more robust to distribution shifts while preserving high accuracy on the target task, with $\alpha=0.5$ being the suggested interpolation value. \cite{wortsman22_modelsoups} proposes to average the parameters of multiple models fine-tuned from a common pre-trained initialization with different hyperparameter values (eg. learning rate, num. epochs, weight decay, label smoothing, data augmentations) which leads to better-performing models. They also propose only averaging the models that benefit performance on a held-out validation set, so-called \emph{GreedySoup} which is their main method. \cite{ilharco2023_task-arithmetic} introduces task vectors, i.e. the difference in parameter space between a fine-tuned model's parameters and those of the pre-trained initialization. They show that adding or subtracting such vectors is akin to learning and forgetting tasks respectively.

\paragraph{Neuron importance and non-uniform merging}
\cite{wortsman22_modelsoups} also proposes learning the $\alpha$ parameter for each model being merged through gradient-based optimization (so-called \emph{LearnedSoup}). They also try learning one $\alpha$ for each layer of each network instead of one for the whole network. \cite{matena_raffel2022_fisher_weighted_averaging} frame model merging as an approximate maximization of the joint posterior likelihood over the different models' parameters. Using a Laplace approximation with each model's Fisher information matrix as the precision matrix for each model's posterior allows them to weigh each parameter independently in the linear interpolation. \cite{dhawan2024fedfish} proposes a similar method but for distributed learning settings and motivates their work by framing the merging as function space aggregation. \cite{yadav2023tiesmerging} propose to diminish interferences when merging models by first keeping only the parameters that changed the most during fine-tuning (trim), then choosing the sign of the merged parameter why a weighted average of the remaining parameters (elect sign) and finally averaging the remaining parameters having the correct sign. 


\cite{yang2024representationsurgerymultitaskmodel} introduce \emph{Representation Surgery}, an unsupervised method aimed at reducing representation bias in model merging by aligning the merged model’s representation with that of the individual models. Unlike NP Merge, which optimizes at the parameter level, Representation Surgery operates in the representation space, providing a complementary approach.

%% file: sections/methods.tex
\section{Methodology}\label{sec:methods}

\subsection{Model Merging}
Let $A$ and $B$ denote two artificial neural networks (ANNs) with the same architecture, for simplicity purposes assume they're standard multi-layered perceptrons (MLPs) with linear layers. The output of the $i$-th layer (out of a total of $L$ layers) of model $M\in\{A, B\}$ in response to a given input $x^M_{i-1}\in\mathbb{R}^{n_{i-1}}$ is given by:
\[x^M_i = \sigma\left(W^M_ix^M_{i-1}+b^M_i\right)\]
Where $\sigma$ is the non-linearity, often a ReLU and $W_i\in\mathbb{R}^{n_i\times n_{i-1}}$, $b_i\in\mathbb{R}^{n_i}$ are that layer's parameters, respectively its weights and bias.

Model merging aims to combine the learned features of both models by first aligning the two models and then aggregating their parameters. This can be seen as the successive application of two functions one for aligning and one for aggregation:
\[A', B' = \mathtt{Align}(A, B)\]
\[C = \mathtt{Agg}(A', B')\]
Where $A', B'$ are the ``aligned'' models and $C=\mathtt{Agg}\circ\mathtt{Align}(A,B)$ is the final merged model. Concretely, the scope of the $\mathtt{Align}$ function is to make $A$ and $B$ linearly mode connected, it is implicitly understood that this operation also aligns the representation spaces of both models \cite{neyshabur2020_transfer}. It is common practice to keep one of the models, say $A$, as being the same through the alignment, i.e. $A'=A$, and aligning the second model to it. Most model alignment methods have been motivated by ANN's invariance to permutations and have therefore looked for permutations of the neurons of model $B$ to align them ``optimally'' to those of model $A$. This is done in a layer-by-layer fashion.
Specifically, we are looking for permutations $P_i\in\mathbb{R}^{n_i\times n_i}$ to align the neurons at layer $i$ of model $B$ to those of model $A$. Since the layers are interconnected, we also need to apply the inverse permutation $P^{-1}_i$ at the input level of the following layer's parameters. After alignment, the output of layer $i$ in model $B'$ becomes:

\[x^{B'}_i = \sigma\left(P_iW^M_iP^{-1}_{i-1}x^{B'}_{i-1}+P_ib^M_i\right)\]

We note that some methods go beyond permutations as this methodology can be roughly extended to invertible linear transformations $T_i\in\mathbb{R}^{n_i\times n_i}$ and their inverses $T_i^{-1}$ \cite{horoi2024harmony}. However, we note that ANN non-linearities might interfere with these merge/un-merge transformations in cases where they're not simple permutations.

Previous work on model merging has mainly focused on model alignment, we go over existing methods in Sec. \ref{sec:rw}. Indeed, in most cases, once the models have been aligned, aggregating their parameters has simply been done through uniform linear combinations of the model's parameters, i.e. if $W^{B'}_i= P_i W^{B}_i P^{-1}_{i-1}$ are the weights at layer $i$ of model $B$ after being aligned to model $A$, the weights of the merged model are simply:

\begin{equation}\label{eq:model_interpolation}
    W_i = \alpha W^{A}_i + (1-\alpha)W^{B'}_i
\end{equation}

The merging parameter, $\alpha\in(0,1)$, is a single scalar for the whole merging operation, i.e. a fixed value of $\alpha$ is used for all parameters and all layers being merged, hence the term ``uniform''. The most commonly used value is $\alpha=0.5$.

\subsection{Non-uniform Parameter-wise Model Merging}\label{ss:npmerge}
While this simple model aggregation function works relatively well in practice, it assigns equal importance to all parameters of the models being merged which might not be an appropriate assumption. Especially in situations where the models were trained on disjoint datasets, it is natural to think that one of the two models might have learned certain features to a greater or lesser extent than the other model. Therefore, assigning equal importance to all parameters by fixing $\alpha=0.5$ might be detrimental to the model being merged. Instead, we argue it might be beneficial to assign a different weight to each parameter. Mathematically this means that instead of having $\alpha$ be a scalar (shared by all layers), we would have for each layer $i$ a tensor $\boldsymbol{\alpha}_i\in\mathbb{R}^{n_i\times n_i}$ of the same size as the models weight tensors at that layer $W_i^M\in\mathbb{R}^{n_i\times n_i}$. The parameters of the merged models then become:

\begin{equation}\label{eq:non-uniform_param_interp}
W_i = \boldsymbol{\alpha}_i \odot W^{A}_i + (\mathbf{1}-\boldsymbol{\alpha}_i)\odot W^{B'}_i
\end{equation}

\noindent Where $\odot$ is the elementwise product and $\mathbf{1}$ is the matrix full of ones of the necessary size. Another way to visualize this is that we have an $\boldsymbol{\alpha} = \bigcup_{i=1} \boldsymbol{\alpha}_i$ tensor, the same size as one of the two models' parameters, and for each parameter from the original models we have a scalar in $\boldsymbol{\alpha}$ which corresponds to its linear interpolation weight. The question then naturally becomes what weight should we assign to each parameter? We propose to automate this process by \emph{learning} the elements of $\boldsymbol{\alpha}$ through gradient-based optimization.


\subsection{Extending NP Merge to the multi-model setting}
NP Merge naturally works in the setting where two models are merged by optimizing the interpolation values ($\alpha$'s) between the two models. If we merge two models, each having $n$ parameters this means we are optimizing $n$ alphas. In the multi-model setting, if we had $m$ models each with $n$ parameters then we would need to optimize simultaneously $(m-1)\times n$ alphas, which can quickly become overwhelming regarding memory usage and compute. Instead, we extend NP Merge to be able to merge more than two models by doing successive pairwise merges. In other words, we start by merging pairs of models chosen at random from the ones to be merged using NP Merge, then we similarly merge the resulting models until we get a final unique model.



\subsection{Comparison Between NP Merge and Fine-Tuning}\label{ss:npmerge_v_ft}

Since NP Merge uses labeled data to optimize the interpolation parameters through gradient descent-based training, and since, for a given merge, the number of scalar interpolation parameters being optimized is the same as the number of parameters inside one of the models, it is natural to compare NP Merge to standard fine-tuning of the model post merging.

A key distinction between NP Merge and standard fine-tuning is that in NP Merge we constrain the $\alpha$ values to be in the range $(0,1)$. In practice, this is implemented using a sigmoid function $\sigma(x) = \frac{1}{1+e^{-x}}$ and by optimizing $\alpha_{pre} \in\mathbb{R}$ such that $\sigma(\alpha_{pre}) = \alpha \in (0,1)$. While it is the $\alpha= \sigma(\alpha_{pre})$ that is used to obtain the final model, it is the $\alpha_{pre}$ that is learned during optimization. This has two important effects:

\textbf{1)} Unlike fine-tuning, where the optimization takes place in the full weight space, NP Merge operates within a \textit{restricted optimization space} by constraining the $\alpha$ values between $0$ and $1$. This ensures that the interpolation remains a combination of the models being merged, rather than diverging too far from the original weights. The $\alpha$ values can be seen as coefficients that balance the contribution of each model at a per-parameter level, effectively allowing the method to adapt to specific differences between the models. This bounded optimization 
%
%
leads to more robust merged models and minimizes potential performance degradation that could arise from overfitting or drastic weight adjustments in fine-tuning.

\textbf{2)} Learning $\alpha_{pre} \in\mathbb{R}$ through the sigmoid function introduces an implicit bias in the optimization process. Indeed, for large magnitude values of $\alpha_{pre}$ (far from 0), the gradient through the sigmoid function is very small since the sigmoid function at those values plateaus. This discourages the values of $\alpha_{pre}$ from staying too far from 0 and keeps the associated $\alpha = \sigma(\alpha_{pre})$ values relatively close to 0.5, i.e. the merged model is kept relatively close to the average of the models being merged.
This constitutes a concrete change in the dynamics of the training procedure for fine-tuning, where the weight updates are unrestricted and can stray far from the initialization value. This produces a regularization effect: models obtained through NP Merge become less prone to overfitting than fine-tuned models.

NP Merge therefore has some of the flexibility of full fine-tuning but is more robust and better adapted to the problem of model merging.

%% file: sections/results.tex
\section{Experiments}
\label{sec:results}
In this section, we experimentally validate the proposed model merging technique. We consider standard model merging settings with similar and dissimilar training data distributions for the base models \cite{horoi2024harmony, stoica2023zipit} as well as a setting where many models are merged.

\subsection{Experimental details}
\label{subsec:experimental_details}
In practical terms, we have implemented Pytorch \cite{paske2019pytorch} models whose weights are given by Eq. \ref{eq:non-uniform_param_interp}. The weights being merged, i.e. those of model $A$ ($W^A_i$) and those of aligned model $B$ ($W^{B'}_i$) are kept constant for all $L$ layers. We can then backpropagate through these layers, compute the training loss gradient for $\boldsymbol{\alpha}$, and update them through gradient descent.
We consider VGG11 \cite{simonyan2015a_vgg} and ResNet20 models \cite{he2016_resnet} of different widths \cite{zagoruyko16_wideresnets} trained on CIFAR10 and CIFAR100 respectively. We also train pairs of ResNet20 models of different widths on three disjoint splits of CIFAR100:
\begin{enumerate}
    \item \textbf{80\%-20\% Split:} one model is trained on 80\% of the data from the first 50 classes and 20\% of the other 50 classes, the second model is trained on the remaining examples, this split was considered in \cite{ainsworth2023_git-rebasin, jordan2023repair};
    \item \textbf{Dirichlet:} we sample from a Dirichlet distribution with parameter vector $(0.5, 0.5)$ to split the training examples of each class into 2 disjoint sets, each class has a different split and each model is trained on one of the two sets;
\end{enumerate}
We always evaluate and report the accuracy of the full test set of the task considered.
Since NP Merge's optimization process involves training the $\mathbf{\alpha}$ parameters (of the same size as any
given neural architecture), we report accuracies after 10 optimization epochs over the data. 
We use the ADAM \cite{KingBa15} optimizer with a learning rate of 0.01 for CIFAR10 and CIFAR100. 
To avoid biasing the results towards any particular model, we initialized the
$\alpha$ parameters to 0.5 for all experiments.
We compare the performance of our NP Merge method with the following:
\begin{itemize}
    \item \textbf{Base models avg.:} the average of the models being merged individual accuracies;
    \item \textbf{Ensemble:} the accuracy of the ensemble made up by the models being merged, here we average the logits of the different models and take the $\mathtt{argmax}$ of the average as the predicted class;
    \item \textbf{Direct averaging:} directly averaging the models without any alignment;
    \item \textbf{Permute:} aligning representation spaces by permuting model activations, where the permutation matrix is found by solving the linear sum assignment problem for maximizing the sum of correlations between paired neurons \cite{li2015_convergent, tatro2020_opt-mode_con}. The aggregation of the aligned models' parameters is done by simply averaging;
    \item \textbf{Weight Matching:} aligning representation spaces by permuting each layer parameter; the permutation matrix is
    found by solving the sum of bilinear alignment problems for maximizing the correlations between paired neurons \cite{ainsworth2023_git-rebasin}.
    Like for \emph{permute}, the aggregated parameters are computed by averaging the aligned models;
    \item \textbf{CCA Merge:} aligning model representation spaces by using CCA to find the optimal linear combination of model $B$'s features maximally correlated to model $A$'s features. This is the method proposed by \cite{horoi2024harmony} and represents the state of the art in this setting of aligning models trained from different initializations that do not share a part of their training trajectories. Here as well aggregation of the aligned models' parameters is done through simple averaging, i.e. $\alpha=0.5$ for all parameters of all models.
\end{itemize}
As discussed in Sec. \ref{ss:npmerge_v_ft}, it is natural to compare our NP Merge method to full fine-tuning of the model post merging. Therefore we also compare NP Merge with standard fine-tuning of the merged model $\frac{A + B'}{2}$, using the same data for both methods.

We note that since NP Merge is only concerned with the aggregation of parameters part of model merging, the alignment of the representation spaces needs to be done with one of the existing alignment methods. In all reported results for NP Merge we align the models using the permutations provided by the Permute method \cite{tatro2020_opt-mode_con} and then do the proposed gradient-based optimization of the interpolation values for the parameter aggregation. We reset the Batch Norm statistics post merging to avoid variance collapse as suggested by \cite{jordan2023repair}.

\subsection{Merging models trained on the same data distribution}\label{ss:same_distribution}
Table \ref{tab:results_same_init} present the test
accuracies and standard deviations of the different merging methods considered for VGG11 and ResNet20 models trained on CIFAR10 and
CIFAR100 respectively.

    
\begin{table}[ht]
\renewcommand{\arraystretch}{1.3}
  \caption{
    ResNet results and comparisons with baselines.
  }
  \centering
    \begin{tabular}{l c c}
    \Xhline{0.75pt} 
    & CIFAR-10 & CIFAR-100 \\
    & VGG11$\times 1$ & ResNet20$\times 1$ \\
    \hline
    \textcolor{gray}{Base models avg.}& \textcolor{gray}{87.27 \small{$\pm$ 0.25\%}}  &
    \textcolor{gray}{69.21 \small{$\pm$ 0.22\%}} \\ 
    \textcolor{gray}{Ensemble}        & \textcolor{gray}{89.65 \small{$\pm$ 0.13\%}} &
    \textcolor{gray}{73.51 \small{$\pm$ 0.20\%}} \\
    \textcolor{gray}{Direct averaging}& \textcolor{gray}{10.54 \small{$\pm$ 0.93\%}} &
    \textcolor{gray}{1.61 \small{$\pm$ 0.16\%}}\\
    \hline
    Permute          & 54.39 $\pm$ 6.45\% & 28.76 $\pm$ 2.20\% \\
    Weight Matching    &  80.12 $\pm$ 1.91\% & 18.00 $\pm$ 1.61\% \\
    CCA Merge & 82.65 $\pm$ 0.73\% & 31.79 $\pm$ 1.97\% \\
    \hline
    Finetune-P & 88.11\% $\pm$ 0.21\% & 39.37\% $\pm$ 1.74\% \\
    Finetune-WM & 88.54\% $\pm$ 0.03\% & 32.98\% $\pm$ 0.66\% \\
    \hline
    \textbf{NP-P} & \emph{88.38\% $\pm$ 0.23\%} & \textbf{\emph{62.88\% $\pm$ 0.11\%}} \\
    \textbf{NP-WM} & \textbf{\emph{88.49\% $\pm$ 0.16\%}} & \emph{61.68\% $\pm$ 0.50\%} \\
  \end{tabular} 
  \label{tab:results_same_init}
\end{table}

Our NP Merge method can consistently outperform past model merging methods, as well as, fine-tuning the merged model. Ensembling is considered a strong upper bound for model merging, nevertheless, our method
closes the gap with the ensemble accuracy.

To assess whether NP Merge can be combined with different model alignment techniques, we try using the merged models from both the Permute and the Weight Matching alignment methods as initializations for NP Merge (NP-P and NP-WM respectively).
Both merging priors were able to beat the other strong baselines; moreover, we observed that
NP Merge yielded performance gains over fine-tuning the merged model, especially in the more complex setting with ResNet models trained on CIFAR100. Finally, we have used the complete CIFAR training dataset for either fine-tuning
or NP Merging.


\subsection{Merging models trained on different data distributions}
\label{ss:results-nonuniform}

To assess the performance of our method in harder scenarios, we conducted experiments for merging models trained on differing data
distributions, as described in Section \ref{subsec:experimental_details}. The results in Tables 
\ref{tab:cifar100_results} and \ref{tab:imagenet200_results} show that NP Merge can outperform all the baseline accuracies, including the 
ensemble accuracy, on all the non-uniform class distributions. For both sets of results, the $\alpha$ parameters were optimized in the same 
way as described in Section \ref{subsec:experimental_details}, i.e., running 10 epochs on the whole training set. 
For Table \ref{tab:imagenet200_results}, particularly, we used the ImageNet-200 dataset, which is restricted to only 200 classes.
Moreover, our results show that aligning models via a permute prior implies
a better outcome after the proposed NP-optimization, as demonstrated empirically in both Tables \ref{tab:cifar100_results}
and \ref{tab:imagenet200_results}.

Similarly to the results presented in Table \ref{tab:results_same_init}, we compare NP Merge with 
fine-tuning the merged models. We continue observing that NP Merge can beat fine-tuning when using the same
prior alignment and merging algorithm. Furthermore, we show that the outcome of NP Merge yields similar results when using an activation-based alignment method (Permute) and a weight-based alignment method (Weight Matching).

\begin{table}[ht]
\renewcommand{\arraystretch}{1.3}
    \caption{
      Accuracies and standard deviations for ResNet20$\times$8 on CIFAR-100.
    }
    \centering
    \begin{tabular}{l c c}
        \Xhline{0.75pt} 
        & 80\%-20\% & Dirichlet \\
        \hline
        \textcolor{gray}{Base models avg.}  & \textcolor{gray}{65.66 \small{$\pm$ 0.71\%}} & \textcolor{gray}{59.98 \small{$\pm$ 1.80\%}} \\
        \textcolor{gray}{Ensemble}          & \textcolor{gray}{77.84 \small{$\pm$ 0.23\%}} & \textcolor{gray}{73.77 \small{$\pm$ 0.44\%}} \\
        \textcolor{gray}{Direct averaging}  & \textcolor{gray}{11.40 \small{$\pm$ 1.62\%}} & \textcolor{gray}{20.55 \small{$\pm$ 3.07\%}} \\
        \hline
        Permute         & 62.11 $\pm$ 0.30\% & 58.45 $\pm$ 1.76\% \\
        Weight Matching & 60.86 $\pm$ 0.01\% & 52.20 $\pm$ 3.39\% \\
        CCA Merge       & 66.35 $\pm$ 0.19\% & 60.38 $\pm$ 1.68\% \\
        \hline
        Finetune-P      & 72.50\% $\pm$ 0.03\% & 72.28\% $\pm$ 0.23\% \\
        Finetune-WM     & 71.95\% $\pm$ 0.08\% & 67.96\% $\pm$ 0.90\% \\
        \hline
        \textbf{NP-P}   & \textbf{\emph{73.13\% $\pm$ 0.12\%}} & \textbf{\emph{73.45\% $\pm$ 0.08\%}} \\
        \textbf{NP-WM}  & \emph{72.94\%} $\pm$ 0.01\% & \textbf{\emph{73.45\% $\pm$ 0.18\%}} \\
    \end{tabular}
\label{tab:cifar100_results}
\end{table}

\begin{table}[ht]
    \renewcommand{\arraystretch}{1.3}
    \caption{
      Accuracies for ResNet18$\times$4 on ImageNet-200 (Dirichlet setting), including Top1\% and Top5\% accuracies.
    }
    \centering
    \begin{tabular}{l c c}
        \Xhline{0.75pt} 
        & \multicolumn{2}{c}{Dirichlet} \\
        & Top1\% Acc. & Top5\% Acc. \\
        \hline
        \textcolor{gray}{Base models avg.}  & \textcolor{gray}{66.67 \small{$\pm$ 0.54\%}} & \textcolor{gray}{85.77 \small{$\pm$ 0.54\%}} \\
        \textcolor{gray}{Ensemble}          & \textcolor{gray}{75.96 \small{$\pm$ 0.85\%}} & \textcolor{gray}{92.88 \small{$\pm$ 0.38\%}} \\
        \textcolor{gray}{Direct averaging}  & \textcolor{gray}{0.06 \small{$\pm$ 0.03\%}} & \textcolor{gray}{0.24 \small{$\pm$ 0.09\%}} \\
        \hline
        Permute         & 51.53 $\pm$ 1.04\% & 78.40 $\pm$ 1.07\% \\
        Weight Matching & 30.74 $\pm$ 0.21\% & 55.28 $\pm$ 0.60\% \\
        CCA Merge       & 58.85 $\pm$ 1.20\% & 81.33 $\pm$ 0.78\% \\
        \hline
        Finetune-P      & 62.59 $\pm$ 0.44\% & 84.79 $\pm$ 0.47\% \\
        Finetune-WM     & 58.88 $\pm$ 0.49\% & 81.47 $\pm$ 0.28\% \\
        \hline
        \textbf{NP-P}   & \textbf{62.21 $\pm$ 2.07\%} & \textbf{86.31 $\pm$ 1.20\%} \\
        \textbf{NP-WM}  & 58.55 $\pm$ 2.08\% & 82.69 $\pm$ 1.12\% \\
    \end{tabular}
\label{tab:imagenet200_results}
\end{table}

\TODO[done]{for each subsection where you describe an experiment (e.g., Table 1, Table 2), add a sentence or two clarifying the size of the optimization dataset used in each case.}

\subsection{How much data to use for NP Merge?}

Many model fusion algorithms, including those mentioned in Section \ref{subsec:experimental_details}, depend on
a suitable dataset as a prerequisite. Methods like Permute and CCA Merge assume such data is available,
yet little has been discussed about its effect on each algorithm's outcomes. For simplicity, we refer to this
dataset as the \emph{optimization dataset}, and we extract different sizes the same way validation data is conventionally
sampled from training data.

Table \ref{tab:results_per_val} breaks down our model fusion methods' performance according to the amount of data available, 
we select $k$ examples per class for $k \in \{100, 10, 5, 1\}$ and use the whole training dataset for optimization. 
For this experiment, we have used ResNet20$\times$8 models on CIFAR-100, following our \emph{unbalanced}
setting. As expected, for every reported method, we observe a decreasing performance pattern as the dataset size decreases;
more importantly, we show that NP remains more stable than fine-tuning even subsets as small as 1\% of the 
training data (5 examples per class). For fine-tuning, we have used the ADAM optimizer and have
found optimal hyper-parameters for all settings. 

Model fusion becomes especially challenging in scenarios where there is not enough data. Yet these settings are
more realistic than assuming that one has access to all needed data: hence merging algorithms that are robust
to small optimization data shall have a greater impact on further applications. Our findings state that simply aligning
parts of models is not enough, that is, finding a linearly connected version of model $B$ on model $A$'s loss basin.
On the other hand, treating model alignment and weight interpolation as independent optimization problems gives benefits
in the most difficult data settings.

\begin{table*}[ht!]
    \renewcommand{\arraystretch}{1.3}
    \caption{
        Impact of optimization data size on NP-Merge and fine-tuning for ResNet20$\times$8 models on CIFAR-100.
    }
    \centering
    \begin{tabular}{c c c c  c c}
        \Xhline{0.75pt} 
        & & \textbf{MERGING} 
        & & \multicolumn{2}{c}{\textbf{MERGE + OPTIMIZE}} \\
        \textbf{Merging Function} & \textbf{Opt. Data} & \textbf{ONLY} &  & \textbf{NP} & \textbf{Finetune} \\
        \hline
        \multirow{5}{*}{\makecell{Weight Matching \\ (after batch norm reset)}} & \emph{train data} 
        & 60.20\% $\pm$ 0.56\% & & \textbf{73.45\% $\pm$ 0.18\%} & 67.96\% $\pm$ 0.90\% \\
        & 100 ex/class & 51.78\% $\pm$ 2.68\% & & \textbf{67.47\% $\pm$ 0.29\%}  & 67.15\% $\pm$ 0.48\% \\
        & 10 ex/class & 51.57\% $\pm$ 2.64\% & & \textbf{60.42\% $\pm$ 1.09\%}  & 58.37\% $\pm$ 1.70\% \\
        & 5 ex/class & 51.51\% $\pm$ 2.51\% & & \textbf{58.40\% $\pm$ 1.49\%}  & 56.78\% $\pm$ 1.86\% \\
        & 1 ex/class & 50.62\% $\pm$ 2.60\% & & \textbf{55.94\% $\pm$ 1.57\%}  & 55.64\% $\pm$ 1.83\% \\
        \hline
        \multirow{5}{*}{Permute} & \emph{train data} & 62.44\% $\pm$ 1.53\% & & \textbf{73.45\% $\pm$ 0.08\%}  & 72.28\% $\pm$ 0.23\% \\
        & 100 ex/class & 58.21\% $\pm$ 0.19\% & & \textbf{68.53\% $\pm$ 0.13\%}  & 67.32\% $\pm$ 0.43\% \\
        & 10 ex/class & 57.67\% $\pm$ 0.98\% & & \textbf{63.05\% $\pm$ 0.33\%}  & 61.90\% $\pm$ 0.45\% \\
        & 5 ex/class & 57.37\% $\pm$ 0.59\% & & \textbf{61.45\% $\pm$ 0.41\%}  & 60.41\% $\pm$ 0.33\% \\
        & 1 ex/class & 54.20\% $\pm$ 0.26\% & & 56.45\% $\pm$ 0.31\% & \textbf{56.97\% $\pm$ 0.31\%}  \\
    \end{tabular}
\label{tab:results_per_val}
\end{table*}


\subsection{Many model merging}\label{ss:many_models}

This section extends our settings to account for more than 2 models. This a significantly more challenging scenario,
as the number of possible merges grows exponentially with the number of models. We conducted experiments with 2, 4, and 8 models.
This scenario finds important applications in distributed learning settings, such as federated learning, where a central server
needs to merge the models from different clients.

To leverage our 2-model merging performance gains into many models, we arbitrarily form model pairs and run our NP
merging $\alpha$ optimization. Indeed, this includes aligning every pair using the Permute method, the
usual procedure we take for our method. Subsequently, every pair's optimized and merged output models will be
paired again with other output models so that this procedure continues until a final model is obtained. 

\begin{figure}[ht!]
    \centering
    \includegraphics[width=0.5\textwidth]{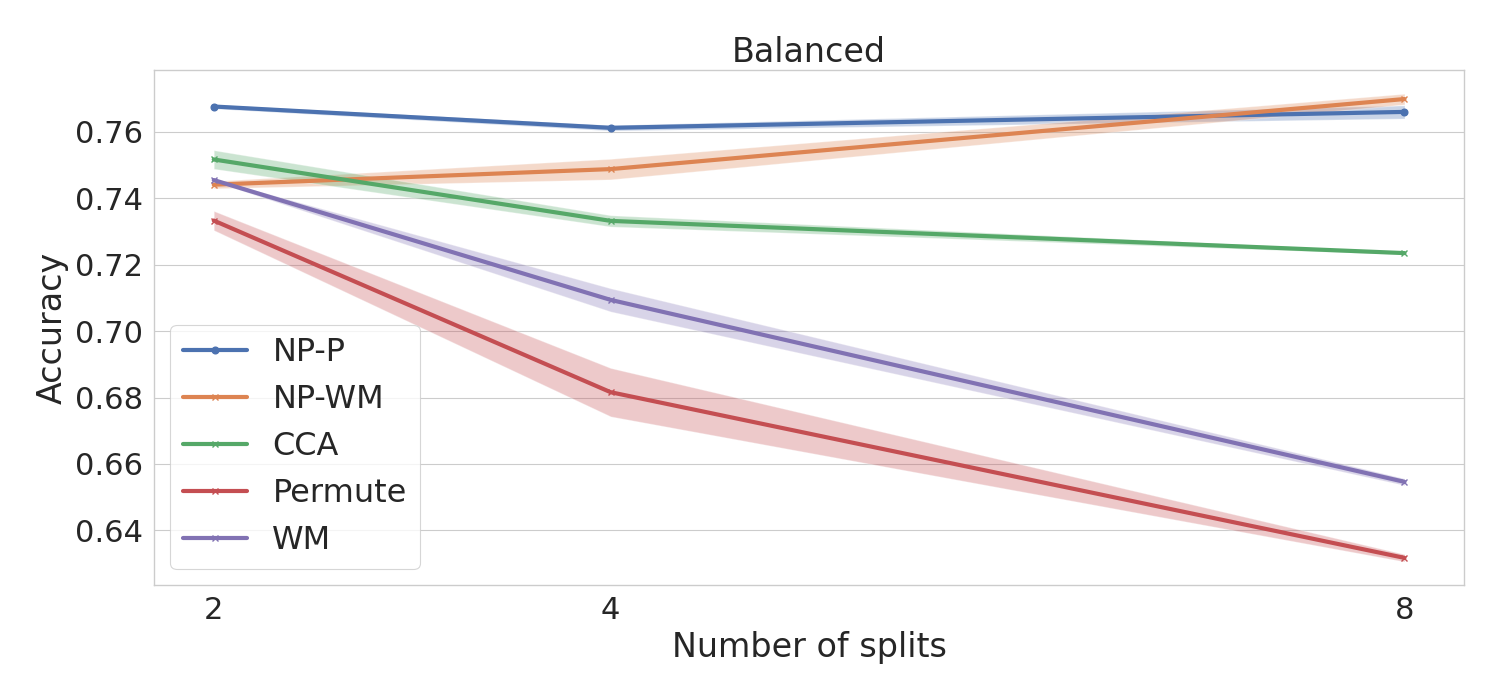}
    \includegraphics[width=0.5\textwidth]{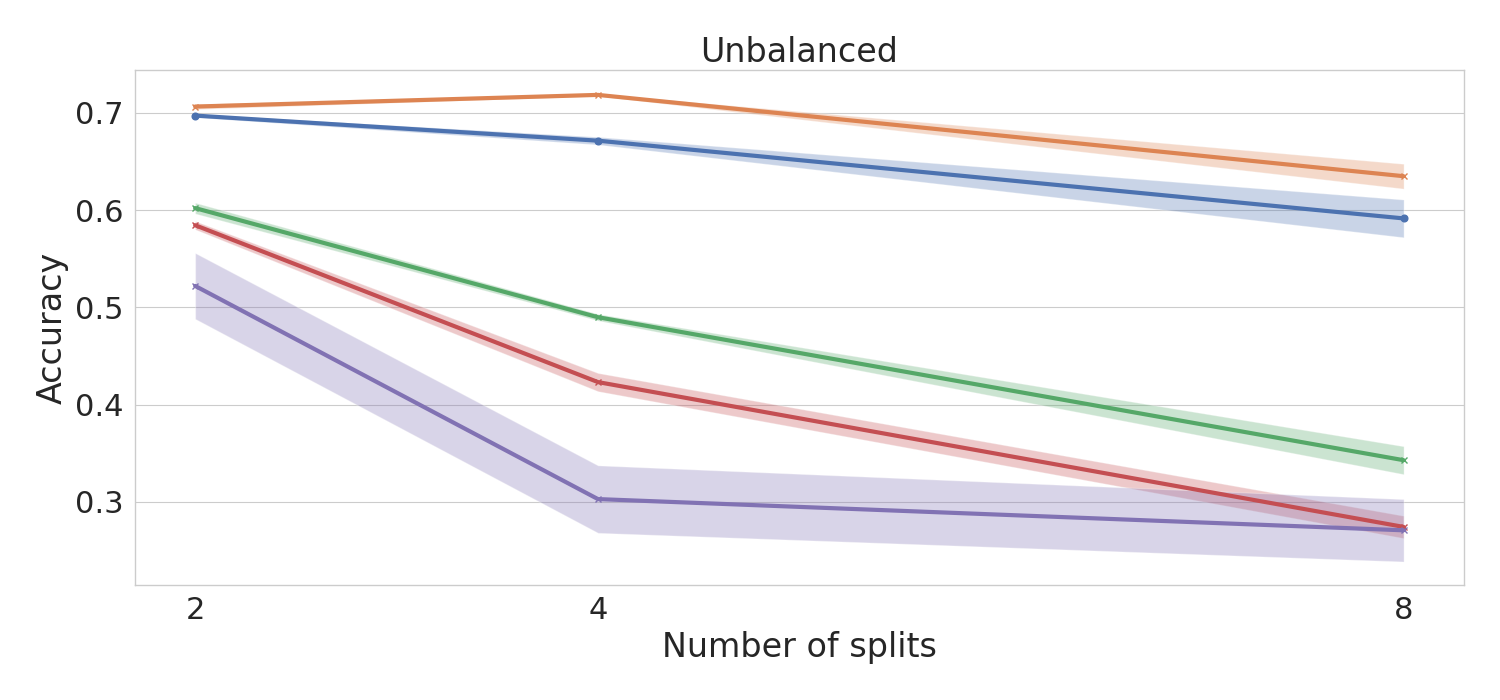}
    \caption{
      We plot the number of merged models against the merged model's accuracy.
      The top plot shows the results for the balanced data setting, and the bottom plot shows the results for the unbalanced data setting.
      This setting corresponds to ResNet20$\times$8 trained on CIFAR100.
      $\alpha$ parameters were optimized in the same way as described in Section \ref{subsec:experimental_details}.\\
    }
    \label{fig:accuracy_vs_num_models}
\end{figure}

In Figure \ref{fig:accuracy_vs_num_models} we compare our method with two baselines: CCA and Permute merge. 
These baselines have been run in a \emph{all-to-one} fashion, where an arbitrary model was chosen as a reference
for which the other ones will be transformed, according to the criteria of each method.
As the number of models increases, the accuracy of the baseline methods decreases; yet in the balanced data setting, where we
merge models initialized equally, our method can maintain a stable accuracy. In the unbalanced data setting, our method still
outperforms the baselines, but the accuracy decreases as the number of models increases. This is expected, as splitting the
dataset into more than 2 models will result in very unbalanced -- and smaller-- sub-datasets. For the unbalanced data setting,
we continue using the Dirichlet distribution on the class labels, as described in Section \ref{subsec:experimental_details}.

Figure \ref{fig:accuracy_vs_num_models} reveals clear trends in performance as the number of models being merged increases. 
Notably, in balanced data scenarios, NP Merge demonstrates stable accuracy, outperforming baseline methods even as the number of models grows. 
This stability suggests that NP Merge effectively handles the increased complexity without significant performance degradation.


This experiment shows a clear divergence from our previous observations, where NP-WM and NP-P usually depicted similar behavior. In the unbalanced case, we particularly observe a dominant performance of the weight matching prior over permute. Both methods indicate a promising path toward aggregating more than 8 models. Furthermore, in the balanced case, both methods demonstrate an increasing trend as the number of models increases; this is the only scenario where we observe such a trend.

\TODO[done]{
    \begin{itemize}
    \item Briefly discuss the computational and memory overhead of NP Merge compared to simpler methods like weight averaging.
    \item Make sure I emphasize model merging fine-grained flexibility
    \end{itemize}
}

\subsection{Memory Trade-offs and Gains}
While NP Merge provides notable performance improvements over traditional model averaging methods, it does come with a certain 
level of computational and memory overhead. This is primarily due to the gradient-based optimization of the interpolation parameters ($\alpha$)
for each weight, which requires additional memory to store these parameters and increases the computation needed to optimize them over multiple iterations.
In contrast, simpler methods such as uniform averaging of model weights or layer-wise interpolation are more computationally efficient 
but lack the fine-grained flexibility that NP Merge offers. Despite these trade-offs, we find that the improvements in accuracy and generalization,
particularly in the more challenging scenarios presented, justify the additional resource requirements.

Our results show that even in the most complex setups, such as the disjoint CIFAR-100 split (as seen in Table \ref{tab:cifar100_results})
NP Merge provides notable accuracy gains over standard methods. Furthermore, Table \ref{tab:imagenet200_results} conveys analogous conclusions,
as we show that adding a computational overhead also benefits our ImageNet-200 results.
These performance gains justify the extra computational effort, particularly 
in high-accuracy environments where precision is prioritized over computational efficiency. While simpler methods may be preferable
in resource-constrained scenarios, NP Merge is ideal for applications where performance is critical, such as in multi-task learning or federated learning.

%% file: sections/discussion_and_conclusion.tex
\section{Discussion and conclusion}
\label{sec:disc_conc}

In recent years, numerous methods have been proposed to align the representation spaces of deep learning models trained from different initializations that do not share parts of their training trajectories. While these methods have helped lower the loss barrier between merged models, achieving linear mode connectivity (LMC) is still challenging in some cases. Instead of focusing solely on model alignment as past works have done, we concentrated on the aggregation of models' parameters after alignment—a research direction that has not received as much attention. Traditionally, merged model parameters are averaged or undergo uniform linear interpolation, which may not fully exploit the models' potential.

Our experiments demonstrate that NP Merge consistently outperforms traditional merging methods and ensemble techniques. NP Merge significantly improves accuracy when merging models trained on the same data distribution and non-uniform class distributions. It also proves robust with limited data, maintaining stable performance with small data subsets. Additionally, NP Merge shows scalability and effectiveness in complex scenarios like federated learning, handling increased computational complexity and memory requirements through successive pairwise merging.

Future work should explore the relationship between learned interpolation values from gradient descent and those from models' Fisher information matrices as suggested in \cite{matena_raffel2022_fisher_weighted_averaging, dhawan2024fedfish}. This line of work must also address the extra computational
overhead issue: it is natural to assume that many distributed applications come with restricted computational budgets. Additionally, investigating the applicability of NP Merge in other domains and with different neural architectures could provide further insights into its versatility and robustness.

\TODO[done]{
    \begin{itemize}
        \item Highlight that despite the computational overhead, NP Merge achieves significant accuracy gains in challenging settings (e.g., disjoint datasets, multi-model merging). Suggests that these gains make NP Merge ideal for high-accuracy scenarios.
        \item discuss addressing memory costs as future work
    \end{itemize}
}